\documentclass[journal]{IEEEtran}
\usepackage{amsmath,amsfonts}
\usepackage{algorithmic}
\usepackage{algorithm}
\usepackage{array}
\usepackage[caption=false,font=normalsize,labelfont=sf,textfont=sf]{subfig}
\usepackage{textcomp}
\usepackage{stfloats}
\usepackage{url}
\usepackage{verbatim}
\usepackage{graphicx}
\usepackage{cite}
\usepackage{multirow}
\usepackage{multicol}
\usepackage{utfsym}
\usepackage{booktabs}
\usepackage{tabularx}
\usepackage[capitalise]{cleveref}
\crefname{figure}{Fig.}{Figs.}
\Crefname{figure}{Fig.}{Figs.}

\usepackage{amsthm}
\newtheorem{theorem}{Theorem}

\begin{document}

\title{Generalizable AI-generated Image Detection \\Based on Fractal Self-similarity in the Spectrum}

\author{Shengpeng Xiao, Yuanfang Guo,~\IEEEmembership{Senior Member,~IEEE}, Heqi Peng, Hui Miao, Zeming Liu, Liang Yang, \\Jiantao Zhou,~\IEEEmembership{Senior Member,~IEEE}, and Yunhong Wang,~\IEEEmembership{Fellow,~IEEE}

\thanks{This work was supported in part by the National Natural Science Foundation of China under Grant 62272020, in part by the Hebei Natural Science Foundation (No. F2024202047), in part by State Key Laboratory of Complex \& Critical Software Environment (SKLCCSE-2025ZX-23), and in part by the Fundamental Research Funds for Central Universities.
\textit{(Corresponding author: Yuanfang Guo. email: andyguo@buaa.edu.cn)}}

\thanks{Shengpeng Xiao, Yuanfang Guo, Heqi Peng, Hui Miao, Zeming Liu, and Yunhong Wang are with the School of Computer Science and Engineering, Beihang University.}

\thanks{Liang Yang is with the School of Artificial Intelligence, Hebei University of Technology.}

\thanks{Jiantao Zhou is with the State Key Laboratory of Internet of Things for Smart City and Department of Computer and Information Science, University of Macau.}
}

\maketitle

\begin{abstract}
With the rapid development of image synthesis techniques, AI-generated images have become increasingly realistic, which heightens the potential risk associated with their misuse and creates a growing need for reliable detection. However, the growing diversity of generative models makes it increasingly difficult for detectors to generalize to images produced by unseen generators. Most existing methods rely on artifacts associated with specific generators, which limits their generalization to images produced by unseen models. To address this problem, we investigate structural characteristics arising from the image generation process itself.  Image generation fundamentally involves constructing spatially rich content from more compact representations, while preserving the semantic identity of structures across different spatial locations. We formalize these properties through dimension-increasing shift-equivariant transformations and show that such transformations induce a self-similar structure in the Fourier spectrum. Across successive generation stages, this structure can propagate recursively and form a hierarchical fractal self-similar pattern. Consequently, different spectral sub-regions exhibit consistent structural correspondences inherited from the generation process, providing a generator-agnostic cue for detection. Based on this observation, we propose Fractal-CNN, which captures spectral self-similarity rather than generator-specific spectral values. Extensive experiments across diverse GAN- and diffusion-based generators demonstrate that Fractal-CNN achieves strong cross-generator generalization, with an average detection accuracy of 93.93\% across 16 test generators.
\end{abstract}


\section{Introduction}
\IEEEPARstart{I}{mage} generation techniques have made significant strides in recent years. Generative models, such as VAE \cite{kingma2013auto}, GAN \cite{goodfellow2014generative}, diffusion model \cite{sohl2015deep}, and their variants \cite{karras2017progressive, karras2019style, brock2018large, zhu2017unpaired, choi2018stargan, park2019semantic, karras2020analyzing, dhariwal2021diffusion, gu2022vector}, can produce highly realistic images from simple text descriptions or other reference images. Despite these remarkable advances, the potential risks associated with the misuse of image generation techniques have also raised significant concerns. In particular, such techniques have been widely exploited for image forgery, ranging from manipulated facial images \cite{rossler2019faceforensics++, 2017deepfakes, 2016faceswap, thies2016face2face, thies2019deferred} to synthetic artistic paintings \cite{bai2020fake}. These synthetic images raise concerns about content authenticity, as they can be easily misused for misinformation and identity impersonation.

\begin{figure}
\centering
\includegraphics[width=\linewidth]{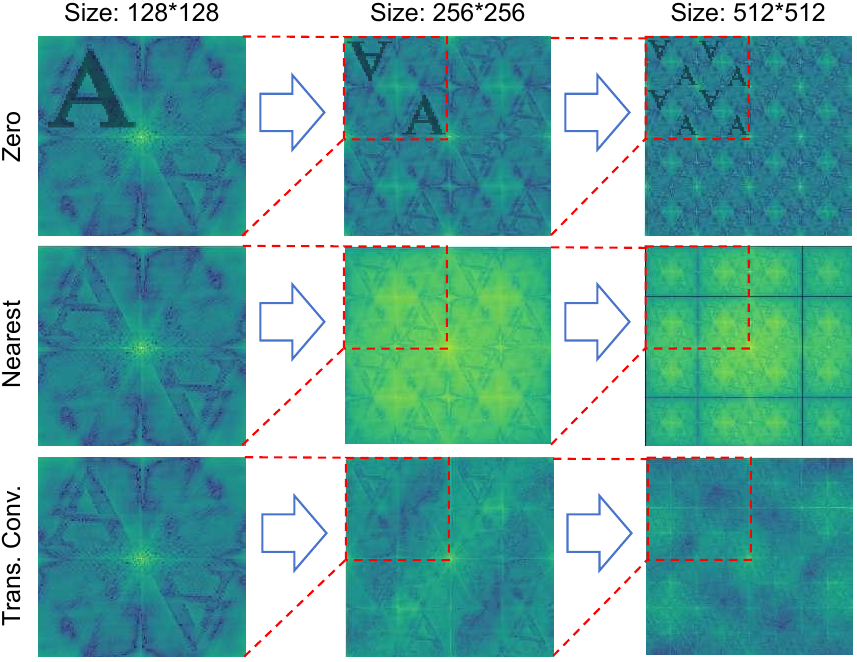}
\caption{Construction of the spectral fractal structure induced by successive dimension-increasing shift-equivariant transformations. Following \cref{eq:fractal_general}, we illustrate the process using zero-filling interpolation, nearest-neighbor interpolation, and transposed convolution. The replicated ``A''-shaped pattern shows that different spectral sub-regions are structurally related through a shared lower-dimensional spectral representation. For better visualization, the spectrum is rearranged such that the low-frequency components are located at the center.}
\label{fig:motivation}
\end{figure}   

Under such circumstances, numerous approaches for detecting AI-generated images have been proposed~\cite{guo2018fake, marra2019gans, liu2020global, wang2020cnn, qi2022principled, ju2022fusing, ojha2023towards, tan2023deepfake, zhong2023patchcraft, liu2022detecting, wang2023dire, zhou2025aigi, miao2025multi, zhong2026self} to mitigate the threats posed by synthetic images. Some methods leverage prior knowledge of specific image content. For example, deepfake detection methods focus on identifying manipulated facial images by detecting anomalous facial attributes or inconsistencies in facial regions \cite{guo2022eyes, matern2019exploiting, tan2023deepfake}. However, such approaches are typically limited to specific image categories (e.g., facial images) and rely heavily on prior knowledge. To improve generalization across diverse semantic content, many studies have instead focused on detecting content-agnostic low-level artifacts in AI-generated images \cite{marra2019gans, zhang2019detecting, frank2020leveraging, durall2020watch, bammey2023synthbuster, corvi2023intriguing, cozzolino2024zero, ricker2024aeroblade, yan2025sanity}. Nevertheless, these artifacts are usually closely related to specific generative models. As generation techniques continue to evolve rapidly, generalizing to images produced by unseen generators has become a critical challenge. To address this issue, recent methods attempt to learn generator-agnostic features of AI-generated images to enhance cross-generator generalization \cite{zhong2023patchcraft, yan2025sanity, zhou2025aigi, cozzolino2024zero, zhang2025detecting}.

Unfortunately, despite the impressive empirical performance achieved by existing detectors, the fundamental characteristics that distinguish AI-generated images from real images remain insufficiently understood. As the diversity of generative methods continues to grow, a deeper understanding of these intrinsic features becomes increasingly important for developing detection methods which can reliably generalize to images produced by unseen models.

In this paper, we conduct a theoretical investigation into the structural characteristics arising from the image generation process itself. We observe that image generation fundamentally involves constructing spatially rich content from more compact representations, while the semantic identity of a structure should remain unchanged across spatial locations. We formalize these properties through dimension-increasing shift-equivariant transformations and theoretically show that such transformations induce a self-similar structure in the Fourier spectrum. When such transformations occur successively across multiple generation
stages, the self-similar structure can propagate recursively across scales, forming a hierarchical fractal pattern. As illustrated in \cref{fig:motivation}, corresponding spectral sub-regions share common
underlying spectral components, providing a generator-agnostic structural characteristic.

Based on this observation, we propose Fractal-CNN to exploit spectral self-similarity as a generator-agnostic cue for AI-generated image detection. Rather than relying directly on generator-specific spectral artifacts, Fractal-CNN characterizes the structural relationships among spectral sub-regions through multi-level fractal units. Extensive experiments across diverse GAN- and diffusion-based generators demonstrate strong cross-generator generalization, achieving an average detection accuracy of 93.93\% across 16 test generators.

Our main contributions are summarized as follows:

\begin{itemize}

\item We theoretically show that dimension-increasing shift-equivariant transformations in the image generation process induce self-similar structures in the Fourier spectrum. Successive such transformations recursively propagate this structure across scales, forming a hierarchical fractal pattern.

\item We identify spectral self-similarity as a generator-agnostic
structural cue. By characterizing relationships among spectral
sub-regions rather than their raw values, the proposed representation
reduces dependence on generator-specific artifacts and improves
generalization to unseen generators.

\item We propose Fractal-CNN, which employs multi-level fractal units to capture spectral self-similarity. Extensive experiments demonstrate strong cross-generator generalization across diverse GAN- and diffusion-based models, achieving an average accuracy of 93.93\% across 16 test generators.

\end{itemize}

\section{Related Work}

\subsection{Image Generation}
Image generation aims to synthesize visually realistic images from compact representations, such as random noise or textual descriptions. Early frameworks primarily relied on Generative Adversarial Networks (GANs)~\cite{goodfellow2014generative}, which learn image distributions through adversarial training. Subsequent works improved image quality and training stability through architectural innovations, including progressive growing~\cite{karras2017progressive}, style modulation~\cite{karras2019style, karras2020analyzing}, and large-scale training strategies~\cite{brock2018large}. More recently, diffusion models have emerged as the dominant paradigm. These models generate high-fidelity images by iteratively denoising Gaussian noise~\cite{ho2020denoising, dhariwal2021diffusion, nichol2021glide}. To enhance efficiency, latent diffusion models (LDMs) execute the diffusion process within a learned latent space~\cite{rombach2022high, ramesh2022hierarchical} with the assistance of Variational Auto-encoders (VAEs)~\cite{kingma2013auto}. Alongside these advancements, the increasingly complex architectures render synthetic images less distinguishable from real ones, which poses significant security risks.

\subsection{AI-Generated Image Detection}
AI-generated image detection aims to distinguish synthetic content from real images by exploiting characteristic artifacts introduced during the generation process.
Early efforts primarily leveraged high-level semantic inconsistencies, such as lighting irregularities~\cite{matern2019exploiting} or abnormal facial structures~\cite{guo2022eyes}. However, these approaches are typically category-specific and fail to generalize across diverse content.
To address this, subsequent research shifted toward low-level statistical cues, which are inherently more model-agnostic. Techniques involving noise pattern analysis~\cite{marra2019gans}, spectral or texture-based CNNs~\cite{wang2020cnn, liu2020global}, and frequency-domain artifacts~\cite{frank2020leveraging} demonstrated initial success in generalizing across different GAN architectures. With the advent of diffusion-based generators, recent studies have further explored the intrinsic statistical properties of generated images. Several methods exploit noise residuals~\cite{liu2022detecting}, reconstruction errors~\cite{wang2023dire, ricker2024aeroblade}, or the fusion of global and local cues~\cite{ju2022fusing}. Notably, recent works reveal that diffusion models introduce distinctive spectral autocorrelation patterns in noise residuals~\cite{corvi2023intriguing, bammey2023synthbuster}, suggesting that correlation-based statistics provide robust detection cues. Related efforts have also investigated pixel correlation differences~\cite{zhong2023patchcraft, yan2025sanity}, entropy-based criteria~\cite{cozzolino2024zero}, and spectral reconstruction similarity~\cite{karageorgiou2025any} to enhance adaptability to unseen generators.

However, most current detectors still rely on model-specific statistical artifacts. This dependence limits their generalization across diverse generative paradigms. While various empirical strategies have been explored to bridge this gap, they often lack a formal theoretical foundation. This fundamental deficiency remains a major bottleneck and motivates a deeper investigation into the invariant properties of the generative process.

\begin{figure*}
\centering
\includegraphics[width=\linewidth]{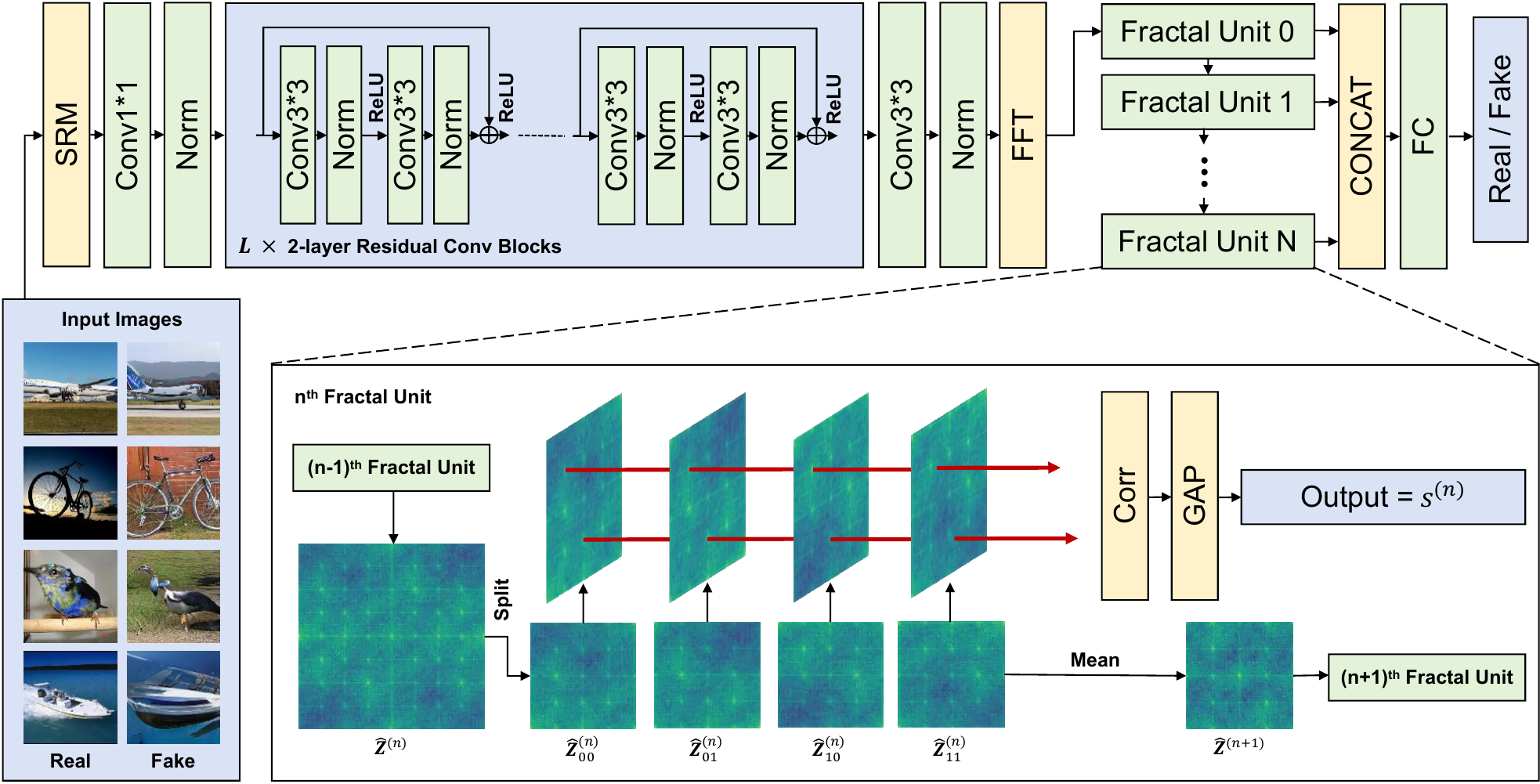}
\caption{Overall architecture of the proposed Fractal-CNN.
Fractal-CNN first learns multiple feature representations to accommodate generator-dependent spectral responses, and then employs multi-level Fractal Units to characterize cross-sub-band self-similarity for AI-generated image detection.}
\label{fig:framework}
\end{figure*}

\section{Methodology}

In this section, we first establish the spectral fractal structure induced by dimension-increasing shift-equivariant transformations. Then we analyze the relationships preserved within this structure across different generators and identify fractal self-similarity as a generalizable detection cue. At last, we introduce Fractal-CNN to characterize this self-similarity for AI-generated image detection.

\subsection{The Fractal Structure of AI-generated Images}

To achieve generalizable detection across different generative models, we seek structural characteristics arising from the generation process itself rather than artifacts tied to specific architectures. A common property of image generation is that compact representations are progressively transformed into spatially richer and higher-dimensional representations. Meanwhile, the identity of a semantic structure should remain unchanged when it appears at a different spatial location. Accordingly, the generation of the same structure at different positions is expected to respond consistently to spatial translations. These two properties motivate us to investigate dimension-increasing, shift-equivariant transformations in the generation process.

Such transformations are common in convolutional and U-Net architectures and in many latent DiT pipelines. Convolutional and U-Net architectures increase spatial resolution through upsampling or decoding stages \cite{karras2017progressive, ho2020denoising}, while many DiT-based models operate on compact latent representations that are subsequently decoded into high-dimensional images \cite{rombach2022high}. Previous studies have shown that these resolution-increasing operations can introduce spectral artifacts \cite{durall2020watch} and structural artifacts in local pixel relationships \cite{tan2024rethinking}. We further investigate the general spectral structure induced by dimension-increasing shift-equivariant transformations, as formalized in \cref{thm:fractal}.

\begin{theorem}
\label{thm:fractal}

Consider an image generation process $\mathcal{G}$ that contains a dimension-increasing shift-equivariant transformation $\Phi: \mathcal{Z} \rightarrow \mathcal{X}$, where $\mathcal{Z} \subset \mathbb{R}^{M\times N}$ and $\mathcal{X} \subset \mathbb{R}^{kM\times kN}$ denote its input and output spatial representation spaces, respectively, and $k>1$ denotes the dimension expansion factor. For any input representation $z \in \mathcal{Z}$, let $x=\Phi(z)$. Then, the spectrum of $x$ exhibits a replicated spectral structure defined by
\begin{equation}
\mathcal{F}(x) =
\mathrm{GFRF}\left(
\left[
\begin{matrix}
\mathcal{F}(z) & \cdots & \mathcal{F}(z)\\
\vdots & \ddots & \vdots\\
\mathcal{F}(z) & \cdots & \mathcal{F}(z)
\end{matrix}
\right]_{k\times k}
\right),
\label{eq:self_similarity}
\end{equation}
where $\mathcal{F}(\cdot)$ denotes the 2D Discrete Fourier Transform, and $\mathrm{GFRF}(\cdot)$ represents the Generalized Frequency Response Functions \cite{jing2015frequency}. $\mathrm{GFRF}(\cdot)$ provides a general representation of nonlinear shift-invariant systems, which in the discrete domain is formulated as
\begin{equation}
\begin{aligned}
\mathrm{GFRF}(S)[u,v]
=&\sum_{p=0}^{\infty}
\sum_{u_1+\cdots+u_p=u,v_1+\cdots+v_p=v}\\
&H_p[u_1,v_1,\cdots,u_p,v_p]
\prod_{i=1}^{p} S[u_i,v_i].
\label{eq:GFRF}
\end{aligned}
\end{equation}
where $p$ denotes the order of the nonlinear response, $H_p$ is the corresponding $p$-th-order frequency-domain kernel, and $S$ is the input spectrum.
\end{theorem}

\begin{proof}
We first formalize the shift-equivariance of the dimension-increasing
transformation $\Phi$. Let $\Delta=(\Delta_m,\Delta_n)$ denote a
spatial translation in the input space. For an input representation
$z\in\mathbb{R}^{M\times N}$, the translation operator is defined as
\begin{equation}
(T_{\Delta}z)[m,n]
=
z[m-\Delta_m,n-\Delta_n],
\end{equation}
where $\Delta_m,\Delta_n\in\mathbb{Z}$. Since $\Phi$ increases the
spatial resolution by a factor of $k$, the same relative spatial
translation corresponds to $k\Delta=(k\Delta_m,k\Delta_n)$ in the
output coordinates. For $x=\Phi(z)\in\mathbb{R}^{kM\times kN}$, the
corresponding translation is
\begin{equation}
(T_{k\Delta}x)[m,n]
=
x[m-k\Delta_m,n-k\Delta_n].
\end{equation}

Therefore, the shift-equivariance of $\Phi$ is expressed as
\begin{equation}
\Phi(T_{\Delta}z)
=
T_{k\Delta}\Phi(z),
\label{eq:shift-equivariance}
\end{equation}
which indicates that the transformation responds consistently to the
same input structure across different spatial locations.

We next examine this constraint in the frequency domain. According to
the shifting property of the 2D DFT,
\begin{equation}
\mathcal{F}(T_{\Delta}z)[u,v]
=
\mathcal{F}(z)[u,v]\,
e^{-j2\pi
\left(
\frac{u\Delta_m}{M}
+
\frac{v\Delta_n}{N}
\right)}.
\end{equation}

Let $E_{\Delta}$ denote the corresponding phase matrix, and define
$\Phi_f=\mathcal{F}\circ\Phi\circ\mathcal{F}^{-1}$ as the
frequency-domain representation of $\Phi$. Then,
\cref{eq:shift-equivariance} is equivalently expressed as
\begin{equation}
\Phi_f\big(\mathcal{F}(z)\odot E_{\Delta}\big)
=
\Phi_f\big(\mathcal{F}(z)\big)\odot E_{k\Delta},
\label{eq:spectral_equivariance}
\end{equation}
where $E_{k\Delta}$ denotes the phase matrix induced by the output
translation $k\Delta$.

Since the output dimension is increased by a factor of $k$,
$E_{k\Delta}$ consists of $k\times k$ repetitions of $E_{\Delta}$:
\begin{equation}
E_{k\Delta}
=
\begin{bmatrix}
E_\Delta & \cdots & E_\Delta\\
\vdots & \ddots & \vdots\\
E_\Delta & \cdots & E_\Delta
\end{bmatrix}_{k\times k}.
\label{eq:phase_replication}
\end{equation}

The spectral constraints in \cref{eq:spectral_equivariance,eq:phase_replication} imply that the output spectrum $\mathcal{F}(x)=\Phi_f\big(\mathcal{F}(z)\big)$ can be represented in the form of \cref{eq:self_similarity}. To demonstrate that any admissible solution is mathematically equivalent, we further show that any alternative solution for $\Phi_f$ is mathematically equivalent to \cref{eq:self_similarity}.

To show that any admissible solution is mathematically equivalent to
\cref{eq:self_similarity}, we proceed by contradiction. Assume that there
exists an alternative solution $\Phi_f'$ satisfying the spectral
constraint but not equivalent to the form in \cref{eq:self_similarity}, such
that
\begin{equation}
\Phi_f'\big(\mathcal{F}(z)\big)
=
\mathrm{GFRF}\left(
\begin{bmatrix}
A_{11} & \cdots & A_{1k}\\
\vdots & \ddots & \vdots\\
A_{k1} & \cdots & A_{kk}
\end{bmatrix}_{k\times k}
\right),
\label{eq:alternative}
\end{equation}
where $A_{ij}\in\mathbb{R}^{M\times N}$,
$i,j=1,\ldots,k$, is a function of $\mathcal{F}(z)$.
According to
\cref{eq:spectral_equivariance,eq:phase_replication}, each $A_{ij}$
must satisfy
\begin{equation}
A_{ij}\big[\mathcal{F}(z)\odot E_\Delta\big]
=
A_{ij}\big[\mathcal{F}(z)\big]\odot E_\Delta.
\label{eq:sub_equivariance}
\end{equation}

This indicates that each $A_{ij}$ satisfies the same shift-equivariance
constraint with respect to $\mathcal{F}(z)$ and can therefore be
represented by a GFRF expansion of $\mathcal{F}(z)$. Substituting these
representations into \cref{eq:alternative} yields a composition of GFRF
transformations, which can itself be represented by another GFRF.
Therefore, any admissible solution reduces to the same replicated-input
structural form as \cref{eq:self_similarity}, completing the proof.
\end{proof}

The above analysis establishes the spectral self-similarity induced by
a dimension-increasing shift-equivariant transformation. In practical
generative models, spatial resolution is commonly increased through
multiple stages, typically with $k=2$. Successive transformations
therefore recursively propagate this structure across scales, while
stochastic or non-fractal components are absorbed into $\epsilon_n$:
\begin{equation}
\begin{aligned}
\mathcal{F}(I) &= \mathcal{F}(z_0),\\
&\mathcal{F}(z_n)
=
\mathrm{GFRF}_n
\left[
\begin{matrix}
\mathcal{F}(z_{n+1}) & \mathcal{F}(z_{n+1})\\
\mathcal{F}(z_{n+1}) & \mathcal{F}(z_{n+1})
\end{matrix}
\right]
+\epsilon_n .
\end{aligned}
\label{eq:fractal_general}
\end{equation}
where $z_n$ denotes the intermediate feature representation at the
$n$-th stage.

\cref{fig:motivation} illustrates this recursive spectral structure using zero-filling interpolation, nearest-neighbor interpolation, and transposed convolution as representative dimension-increasing operations. An ``A''-shaped watermark is introduced only for visualization. Its repeated appearance across spectral sub-regions shows how successive shift-equivariant transformations progressively form the hierarchical structure described in \cref{eq:fractal_general}.

\subsection{Generalizable Fractal Self-Similarity Feature}

The previous analysis shows that dimension-increasing shift-equivariant transformations induce a fractal structure in the spectrum, which can propagate recursively across successive generation stages. For generalizable detection, we further investigate what structural relationships are preserved within this fractal pattern when the generative mapping changes. As shown in \cref{eq:fractal_general}, the input to each GFRF consists of four replicas of the same lower-dimensional spectrum. We therefore
analyze how this replicated fractal structure changes after the GFRF mapping, and which structural relationships are preserved despite such changes.

Following \cref{eq:fractal_general}, we denote the replicated
subsequent-stage spectrum by $Z_{n+1}\in\mathbb{C}^{M_n\times N_n}$:
\begin{equation}
Z_{n+1} =
\begin{bmatrix}
    \mathcal{F}(z_{n+1}) & \mathcal{F}(z_{n+1}) \\
    \mathcal{F}(z_{n+1}) & \mathcal{F}(z_{n+1})
\end{bmatrix},
\label{eq:znp1}
\end{equation}
where $\mathcal{F}(z_{n+1})\in
\mathbb{C}^{M_{n+1}\times N_{n+1}}$ denotes the spectrum of the
subsequent generation stage, and $M_n=2M_{n+1}$, $N_n=2N_{n+1}$ for the $k=2$ case.. The spectrum at the $n$-th stage is then given by
$\mathcal{F}(z_n)=\mathrm{GFRF}_n(Z_{n+1})+\epsilon_n$.

To analyze how this replicated fractal structure is transformed by
$\mathrm{GFRF}_n$, we expand the GFRF response according to
\cref{eq:GFRF}. For each output frequency $(u,v)$, the response consists of products of spectral components in $Z_{n+1}$ whose frequencies jointly contribute to $(u,v)$. We index each such product by $i$. Specifically, the $i$-th term corresponds to a $p$-th-order frequency
tuple $\{(u_j,v_j)\}_{j=1}^{p}$ satisfying
$\sum_j u_j=u$ and $\sum_j v_j=v$, and is written as
\begin{equation}
\left\{
\vphantom{
\begin{matrix}
\mathcal{H}_n^i[u,v]\\[2pt]
\mathcal{Z}_{n+1}^i[u,v]
\end{matrix}
}
\begin{aligned}
&\mathcal{H}_n^i[u,v]
= H_{n,p}[u_1,v_1,\ldots,u_p,v_p],\\[2pt]
&\mathcal{Z}_{n+1}^i[u,v]
= \prod_{j=1}^{p} Z_{n+1}[u_j,v_j].
\end{aligned}
\right.
\label{eq:interaction-term}
\end{equation}
where $\mathcal{Z}_{n+1}^i$ denotes the spectral product inherited from the replicated input structure, while $\mathcal{H}_n^i$ determines its contribution to the output response.

With this notation, the spectral response can be compactly expressed as
\begin{equation}
\mathcal{F}(z_n)[u,v]
=
\sum_{i=0}^{\infty}
\mathcal{H}_n^i[u,v]\mathcal{Z}_{n+1}^i[u,v]
+\epsilon_n[u,v],
\label{eq:gfrf-compact}
\end{equation}
where $\epsilon_n[u,v]$ collects the remaining non-fractal spectral
components.

For a base frequency $(u,v)$, where $0\leq u<M_{n+1}$ and $0\leq v<N_{n+1}$, the four corresponding positions in $Z_{n+1}$ are
$(u+rM_{n+1},v+sN_{n+1})$, where $r,s\in\{0,1\}$. Since $Z_{n+1}$ is formed by four identical copies of $\mathcal{F}(z_{n+1})$, these corresponding positions have the same spectral values. Consequently, the spectral products $\mathcal{Z}_{n+1}^i$ constructed from corresponding frequency combinations are also shared across the four sub-regions. In contrast, the associated kernel responses $\mathcal{H}_n^i$ can vary with the spectral position.
\begin{equation}
\mathcal{F}(z_n)=
\sum_{i=0}^\infty
\mathcal{H}_n^i
\odot
\begin{bmatrix}
\mathcal{Z}_{n+1}^i & \mathcal{Z}_{n+1}^i \\
\mathcal{Z}_{n+1}^i & \mathcal{Z}_{n+1}^i
\end{bmatrix}
+ \epsilon_n.
\label{eq:fractal-decomposition}
\end{equation}

\cref{eq:fractal-decomposition} separates the shared structural
components from their specific spectral responses. The four spectral sub-regions share the same underlying components
$\mathcal{Z}_{n+1}^i$, whereas the kernels $\mathcal{H}_n^i$ modulate their responses at different spectral positions. Therefore, although the input fractal structure is replicated, the four output sub-regions need not have identical spectral values after the GFRF mapping. Instead, they retain a shared-origin structural relationship through the common components $\mathcal{Z}_{n+1}^i$.
\begin{figure}
\centering
\includegraphics[width=0.9\linewidth]{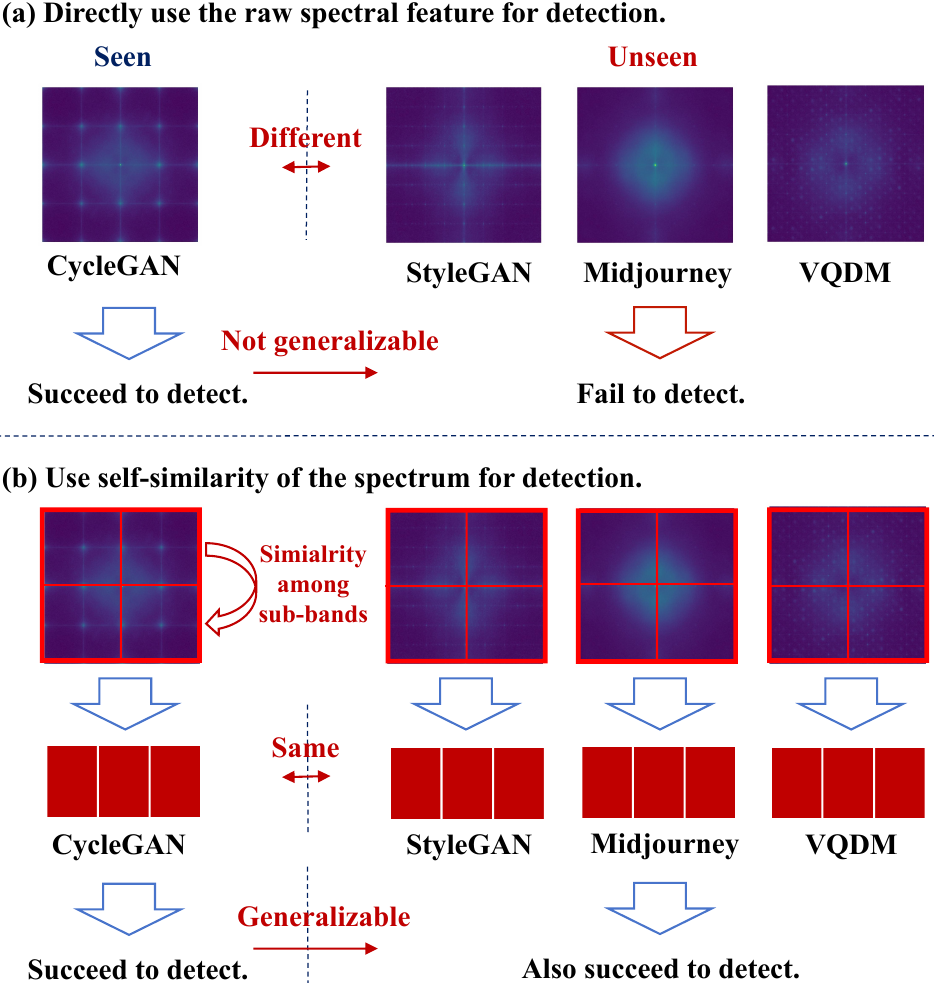}
\caption{Illustration of the generalization mechanism. While raw
spectral features vary across generative models, the structural
self-similarity among corresponding spectral sub-regions remains
comparatively stable.}
\label{fig:generalization-mechanism}
\end{figure}

Because the kernels $\mathcal{H}_n^i$ vary across generators,
the observed spectral responses can also differ substantially.
In contrast, the structural relationship among corresponding spectral sub-regions arises from their shared underlying components $\mathcal{Z}_{n+1}^i$ and thus provides a generator-agnostic structural property. As illustrated in \cref{fig:generalization-mechanism}, this motivates us to characterize the relationships among spectral sub-regions rather than their specific spectral values for generalizable AI-generated image detection.

This observation motivates us to learn feature representations in which the shared-origin relationship among spectral sub-regions can be effectively characterized. Fractal-CNN then explicitly measures this structural self-similarity in the spectral domain, as described in the following subsection.

\subsection{Fractal Convolutional Neural Network}

To exploit the fractal self-similarity characterized in
\cref{eq:fractal-decomposition}, we propose Fractal-CNN.
Since the shared fractal components $\mathcal{Z}_{n+1}^i$ are modulated
by generator-dependent kernels $\mathcal{H}_n^i$ and non-fractal
components $\epsilon_n$, their observed spectral responses can vary
across generators. Fractal-CNN therefore learns multiple feature
representations to accommodate these variations and employs Fractal
Units to characterize cross-region self-similarity at multiple fractal
levels. The resulting multi-level representation $S$ is used for final
classification.

As illustrated in \cref{fig:framework}, Fractal-CNN first applies the Spatial Rich Model (SRM) \cite{fridrich2012rich} to extract diverse local responses from the input image. A convolutional backbone then maps these responses into $C$ feature representations:
\begin{equation}
h=\mathrm{CNN}\left(\mathrm{SRM}(I)\right)
=\left[h_1,h_2,\ldots,h_C\right]
\in\mathbb{R}^{C\times H\times W},
\label{eq:feature-extraction}
\end{equation}
where $h_c$ denotes the $c$-th feature representation. Different generator-dependent kernels
$\mathcal{H}_n^i$ can map the shared fractal components
$\mathcal{Z}_{n+1}^i$ to different observed responses. Instead of
explicitly estimating or inverting these kernels, the backbone learns multiple representations to accommodate such variations, so that the underlying shared fractal components can be effectively characterized across different generators. 

The learned feature representation $h$ is subsequently transformed into the frequency domain to characterize its fractal self-similarity. Its 2D-DFT is computed along the spatial dimensions as
\begin{equation}
\mathcal{F}(h)[u,v]
=
\sum_{x=0}^{H-1}\sum_{y=0}^{W-1}
h[x,y]
e^{-i2\pi\left(\frac{ux}{H}+\frac{vy}{W}\right)}.
\end{equation}
where $\mathcal{F}(h)$ denotes the resulting frequency-domain
representation. Its magnitude $|\mathcal{F}(h)|$ is normalized by a
BatchNorm layer to obtain the level-0 spectral representation
$\hat{Z}^{(0)}$.

The Fractal Units explicitly characterize fractal self-similarity by measuring the structural interactions among corresponding spectral sub-regions at multiple levels. At the $n$-th fractal level, the spectral representation $\hat{Z}^{(n)}$ is split into four spectral sub-regions as
\begin{equation}
\hat{Z}^{(n)}=
\begin{bmatrix}
\hat{Z}^{(n)}_{00} & \hat{Z}^{(n)}_{01} \\
\hat{Z}^{(n)}_{10} & \hat{Z}^{(n)}_{11}
\end{bmatrix}.
\label{eq:fractal=cnn=split}
\end{equation}

Guided by the shared-origin structural prior established in
\cref{eq:fractal-decomposition}, Fractal-CNN characterizes the
relationships among the four spectral sub-regions at each level.
Rather than assuming a fixed numerical similarity among these sub-regions, the feature representation is learned end-to-end so that their cross-sub-band relationships become discriminative for AI-generated image detection. Specifically, the self-similarity feature at the $n$-th level is computed through a four-way cross-sub-band interaction:
\begin{equation}
s^{(n)}=\mathrm{AvgPool}\left(
\hat{Z}^{(n)}_{00}\odot
\hat{Z}^{(n)}_{01}\odot
\hat{Z}^{(n)}_{10}\odot
\hat{Z}^{(n)}_{11}
\right),
\label{eq:similarity-measure}
\end{equation}
where $\odot$ denotes element-wise multiplication and $\mathrm{AvgPool}(\cdot)$ denotes global average pooling. The resulting feature $s^{(n)}$ summarizes the joint responses at
corresponding positions of the four spectral sub-regions and serves as the fractal self-similarity representation at the $n$-th spectral level.

In addition, the recursive spectral structure characterized in
\cref{eq:fractal-decomposition} motivates us to model cross-sub-band interactions at multiple levels. As the corresponding positions in the four sub-regions originate from the same underlying fractal components, we aggregate them through cross-quadrant averaging to construct the representation at the next level as
\begin{equation}
\hat{Z}^{(n+1)}=\frac{1}{4}\left(
\hat{Z}^{(n)}_{00}+
\hat{Z}^{(n)}_{01}+
\hat{Z}^{(n)}_{10}+
\hat{Z}^{(n)}_{11}
\right),
\end{equation}
where $n$ indexes the fractal level. Repeating this process across
$N$ fractal levels yields a set of self-similarity features
$\{s^{(0)},\ldots,s^{(N-1)}\}$. The final multi-level representation $S$ is formed by concatenating these features
\begin{equation}
S=\mathrm{Concat}\left(s^{(0)},\ldots,s^{(N-1)}\right).
\end{equation}

At last, the concatenated self-similarity representation $S$ is input into a fully connected layer for classification. The network is trained end-to-end with the binary cross-entropy loss,
\begin{equation}
\mathcal{L}_{cls}=-\sum_{i}\left[y_i\log(\hat{y}_i)+(1-y_i)\log(1-\hat{y}_i)\right],
\label{eq:loss}
\end{equation}
where $i$ indexes the training samples, $y_i$ denotes the ground-truth label, and $\hat{y}_i$ denotes the predicted probability.

\begin{table*}
\caption{Cross-generator generalization accuracy (\%) on unseen models. All detection models are trained exclusively on ProGAN samples and evaluated on images from diverse, unseen generative architectures to assess zero-shot robustness.}
\label{tab:detection-performance}
\begin{center}
\setlength{\tabcolsep}{1.55mm}{
\begin{tabular}{l| cccc cccc cccc cccc |c}
\toprule
Method & \rotatebox{72}{ProGAN} & \rotatebox{72}{StyleGAN} & \rotatebox{72}{BigGAN} & \rotatebox{72}{CycleGAN} & \rotatebox{72}{StarGAN} & \rotatebox{72}{GauGAN} & \rotatebox{72}{StyleGAN2} & \rotatebox{72}{WFIR} & \rotatebox{72}{ADM} & \rotatebox{72}{Glide} & \rotatebox{72}{Midjourney} & \rotatebox{72}{SDv1.4} & \rotatebox{72}{SDv1.5} & \rotatebox{72}{VQDM} & \rotatebox{72}{Wukong} & \rotatebox{72}{DALLE2} & \rotatebox{72}{Average}\\
\midrule
CNNSpot &\textbf{100.00}&90.17&71.17&87.62&94.60&81.42&86.91&\underline{91.65}&60.39&58.07&51.39&50.57&50.53&56.46&51.03&50.45&70.78\\
FreDect&99.36&78.02&81.97&78.77&94.62&80.57&66.19&50.75&63.42&54.13&45.87&38.79&39.21&77.80&40.30&34.70&64.03\\
Fusing&\textbf{100.00}&85.20&77.40&87.00&97.00&77.00&83.30&66.80&49.00&57.20&52.20&51.00&51.40&55.10&51.70&52.80&68.38\\
LNP&99.67&91.75&77.75&84.10&\underline{99.92}&75.39&94.64&70.85&84.73&80.52&65.55&85.55&85.67&74.46&82.06&88.75&83.84\\
LGrad&99.83&91.08&85.62&86.94&99.27&78.46&85.32&55.70&67.15&66.11&65.35&63.02&63.67&72.99&59.55&65.45&75.34\\
DIRE-G&95.19&83.03&70.12&74.19&95.47&67.79&75.31&58.05&75.78&71.75&58.01&49.74&49.83&53.68&54.46&66.48&68.68\\
DIRE-D&52.75&51.31&49.70&49.58&46.72&51.23&51.72&53.30&\textbf{98.25}&\underline{92.42}&89.45&91.24&91.63&91.90&90.90&92.45&71.53\\
UnivFD&99.81&84.93&95.08&\underline{98.33}&95.75&\textbf{99.47}&74.96&86.90&66.87&62.46&56.13&63.66&63.49&85.31&70.93&50.75&78.43\\
PatchCraft&\textbf{100.00}&92.77&\underline{95.80}&70.17&\textbf{99.97}&71.58&89.55&85.80&82.17&83.79&\underline{90.12}&\textbf{95.38}&\textbf{95.30}&88.91&91.07&\textbf{96.60}&89.31\\
NPR&99.79&\underline{97.70}&84.35&96.10&99.35&82.50&\textbf{98.38}&65.80&69.69&78.36&77.85&78.63&78.89&78.13&76.11&64.90&82.91\\
AIDE&99.99&\textbf{99.64}&83.95&\textbf{98.48}&99.91&73.25&\underline{98.00}&\textbf{94.20}&\underline{93.43}&\textbf{95.09}&77.20&\underline{93.00}&\underline{92.85}&\underline{95.16}&\textbf{93.55}&\textbf{96.60}&\underline{92.77}\\
\midrule 
\textbf{Fractal-CNN}&99.95&95.49&\textbf{97.03}&89.20&99.87&\underline{92.36}&97.29&90.90&91.28&89.38&\textbf{92.95}&92.38&92.47&\textbf{95.34}&\underline{92.99}&94.00&\textbf{93.93}\\
\bottomrule
\end{tabular}
}
\end{center}
\end{table*}

\section{Experiments}

\subsection{Experiment Setups}

\noindent\textbf{Dataset.} The experiments are conducted on the public benchmark AIGCDetectBenchmark \cite{zhong2023patchcraft}. The training set consists of 360,000 fake images from ProGAN \cite{karras2017progressive} and an equivalent number of real images. To evaluate cross-generator detection performance, the test set incorporates samples from 16 diverse generative architectures. These include GAN-based models \cite{karras2017progressive, karras2019style, brock2018large, zhu2017unpaired, choi2018stargan, park2019semantic, karras2020analyzing}, diffusion-based models \cite{dhariwal2021diffusion, rombach2022high, gu2022vector}, and various commercial APIs \cite{Midjourney2023Midjourney, ramesh2022hierarchical, nichol2021glide, jevin2025whichfaceisreal, 2025wukong}. This extensive collection of generative sources ensures a comprehensive assessment of detection robustness against unseen artifacts.

\noindent\textbf{Baselines.} The proposed method is compared to several state-of-the-art detection frameworks, including CNNSpot \cite{wang2020cnn}, FreDect \cite{frank2020leveraging}, Fusing \cite{ju2022fusing}, GramNet \cite{liu2020global}, LNP \cite{liu2022detecting}, LGrad \cite{tan2023learning}, Dire \cite{wang2023dire}, UnivFD \cite{ojha2023towards}, PatchCraft \cite{zhong2023patchcraft}, NPR \cite{tan2024rethinking}, and AIDE \cite{yan2025sanity}. These models provide a comprehensive benchmark for evaluating detection performance across diverse generative sources.

\noindent\textbf{Data augmentations.} To ensure a fair comparison, the data augmentation protocol follows the settings in previous work \cite{yan2025sanity}. The following operations are applied randomly to each sample with a 10\% probability: (1) JPEG compression with a quality factor $\mathrm{QF}\sim\mathrm{Uniform}\{30, ..., 100\}$; (2) Gaussian blur with a standard deviation $\sigma\sim\mathrm{Uniform}[0.1, 3.0]$. After these augmentations, each image is cropped to a resolution of 256$\times$256. For images with dimensions smaller than the target size, padding is applied prior to the cropping operation.

\noindent\textbf{Implementation details.} In our experiments, the SRM kernels are the same as those used in previous works \cite{zhong2023patchcraft, yan2025sanity}. The convolutional backbone consists of a $1\times1$ channel projection followed by $L=8$ residual convolutional blocks. An additional $3\times3$ convolution without activation is applied at the output of the backbone to produce an unconstrained real-valued representation for subsequent spectral analysis. Each convolutional layer utilizes zero padding. Leaky ReLU is employed as the activation function, and batch normalization is applied for feature normalization. During training, the AdamW optimizer is used with a learning rate of $1\times10^{-4}$, while all other parameters follow the default settings in PyTorch. We randomly reserve 10\% of the training data as a validation set. The training process stops when either 20 epochs are reached or the validation loss does not decrease for 5 consecutive epochs. The model corresponding to the lowest validation loss is selected as the final model. The network is trained on 8 NVIDIA 3080 GPUs with a batch size of 32 per GPU.

\subsection{Detection Performance}

In the evaluation, each detection model (except for DIRE-D)is trained exclusively on generated images from ProGAN \cite{karras2017progressive} and real images. The models are subsequently evaluated on real images and generated images from various GAN and diffusion models which are entirely unseen during the training phase. This protocol provides a rigorous assessment of the generalizable detection capability for each method. As reported in Tab. \ref{tab:detection-performance}, our method outperforms existing state-of-the-art approaches in average detection accuracy. Notably, the performance remains stable across diverse generative categories, including GANs, diffusion models, and samples from commercial APIs. The proposed framework achieves an average accuracy of 93.93\%, which establishes a new state-of-the-art result under this challenging setting. Even in the worst case, the detection accuracy remains 89.20\%, which demonstrates substantial robustness to previously unseen generation mechanisms. Notably, our framework maintains superior performance on the WhichFaceIsReal (WFIR) dataset. Despite the complete absence of human face images during the training phase, the model generalizes effectively to this unseen domain. This cross-category success further confirms that the spectral fractal features capture invariant physical patterns rather than dataset-specific semantic content or low-level textures.

\begin{table*}[t]
\centering
\caption{Detection accuracy (\%) under different image perturbations.}
\label{tab:robustness}
\begin{tabular}{lccccccccc}
\toprule
\multirow{2}{*}{Method} & \multirow{2}{*}{Original} & \multicolumn{4}{c}{JPEG Compression} & \multicolumn{4}{c}{Gaussian Blur} \\
\cmidrule(lr){3-6} \cmidrule(lr){7-10}
 &  & QF=95 & QF=90 & QF=75 & QF=50 & $\sigma=1.0$ & $\sigma=2.0$ & $\sigma=3.0$ & $\sigma=4.0$ \\
\midrule
CNNSpot     & 70.78 & 64.03 & 62.26 & 60.65 & 59.66 & 68.39 & 67.26 & 67.13 & 65.85 \\
FreDect     & 64.03 & 66.95 & 67.45 & 66.64 & 65.33 & 65.75 & 66.48 & 68.58 & 69.64 \\
Fusing      & 68.38 & 62.43 & 61.39 & 59.34 & 57.41 & 68.09 & 66.69 & 66.02 & 65.58 \\
LNP         & 83.84 & 53.58 & 54.09 & 53.02 & 52.85 & 67.91 & 66.42 & 66.20 & 62.69 \\
LGrad       & 75.34 & 51.55 & 51.39 & 50.00 & 50.00 & 71.73 & 69.12 & 68.43 & 66.22 \\
DIRE-G      & 68.68 & 66.49 & 66.12 & 65.28 & 64.34 & 64.00 & 63.09 & 62.21 & 61.91 \\
UnivFD      & 78.43 & 74.10 & \underline{74.02} & \underline{69.92} & \underline{68.68} & 70.31 & 68.29 & 64.62 & 61.18 \\
PatchCraft  & 89.31 & 72.48 & 71.41 & 69.43 & 67.78 & 75.99 & 74.90 & 73.53 & 72.28 \\
AIDE        & \underline{92.77} & \textbf{75.54} & \textbf{74.21} & \textbf{70.64} & \textbf{69.60} & \underline{81.88} & \underline{80.35} & \underline{80.05} & \underline{79.86} \\
\midrule
\textbf{Fractal-CNN} & \textbf{93.93} & \underline{75.02} & 72.72 & 68.92 & 64.57 & \textbf{88.07} & \textbf{88.09} & \textbf{87.63} & \textbf{87.41} \\
\bottomrule
\end{tabular}
\end{table*}

In the real world, the perturbations, such as JPEG compression and blurring, usually affect AI-generated image detection. To evaluate the robustness of our model, we conduct experiments under common perturbations, as shown in \cref{tab:robustness}. Under JPEG compression, our method shows a performance trend similar to most of the state-of-the-art detectors. According to \cref{thm:fractal}, the 8$\times$8 block-based quantization can be viewed as a compression-and-regeneration process. This process introduces fractal structure similar to AI-generated images. Therefore, robustness against JPEG compression cannot be achieved through generalizable detection alone in our theoretical framework. It essentially becomes a source attribution task. Generator-agnostic features provide limited discriminative capacity in this case. Meanwhile, Fractal-CNN shows remarkable resilience to Gaussian blur. Our approach maintains stable detection performance across varying blur intensities. Many baseline methods experience noticeable accuracy drops as the parameter $\sigma$ increases. However, Fractal-CNN remains effective by prioritizing structural self-similarity over absolute feature magnitudes. Although Gaussian blur alters the spatial and spectral feature values, it preserves the underlying self-similarity among spectral sub-regions. By leveraging these consistent structural invariants, the proposed model effectively mitigates performance degradation caused by the variation of absolute feature representations.

\begin{table}
\caption{Ablation study concerning the number of Fractal Units (FU), where $N$ denotes the total count of units within the detector. For the configuration with $N=0$, the fractal representation is omitted, and the raw spectrum is utilized directly for classification to assess the baseline performance.}
\label{tab:ablation}
\begin{center}
\begin{tabular}{lcccccccc}
\toprule        
\multirow{2}{*}{Generator} & w/o FU & \multicolumn{4}{c}{w/ FU}     \\
\cmidrule(lr){2-2} \cmidrule(lr){3-6}
& $N=0$   & $N=1$ & $N=2$ & $N=3$ & $N=4$ \\
\midrule
ProGAN    & 99.98 & 99.91 & 99.96 & 99.96 & 99.95 \\
StyleGAN  & 95.96 & 87.34 & 95.93 & 93.96 & 95.49 \\
BigGAN    & 84.93 & 95.92 & 97.40 & 97.05 & 97.03 \\
CycleGAN  & 80.29 & 77.79 & 90.90 & 77.15 & 89.20 \\
StarGAN   & 99.95 & 93.48 & 99.87 & 99.90 & 99.87 \\
GauGAN    & 66.55 & 80.05 & 87.69 & 78.21 & 92.36 \\
StyleGAN2 & 92.31 & 91.28 & 92.21 & 95.00 & 97.29 \\
WFIR      & 63.25 & 77.20 & 83.05 & 77.60 & 90.90 \\
ADM       & 86.81 & 84.97 & 89.62 & 81.46 & 91.28 \\
Glide     & 81.05 & 81.69 & 86.41 & 77.86 & 89.38 \\
Midjourney& 67.00 & 91.06 & 88.86 & 87.43 & 92.95 \\
SDv1.4    & 85.03 & 92.80 & 93.15 & 85.33 & 92.38 \\
SDv1.5    & 84.91 & 92.69 & 93.04 & 93.83 & 92.47 \\
VQDM      & 85.21 & 91.14 & 94.52 & 95.98 & 92.49 \\
Wukong    & 82.10 & 91.74 & 91.43 & 91.53 & 92.99 \\
DALLE2    & 93.80 & 84.70 & 92.80 & 85.05 & 94.00 \\
\midrule
Average   & 84.32 & 88.36 & 92.30 & 89.37 & 93.93 \\
\bottomrule
\end{tabular}
\end{center}
\end{table}

\subsection{Ablation Study}

\noindent\textbf{Effectiveness of the Fractal Units.} To evaluate the contribution of Fractal Units (FU), the number of units in the detector is varied, including the baseline configuration without FU ($N=0$). As shown in \cref{tab:ablation}, the integration of fractal units significantly enhances the overall detection performance. Specifically, the average accuracy increases from 84.32\% to 93.93\%, while the worst-case accuracy improves from 66.55\% to 89.20\%. These results indicate that the exploitation of spectral self-similarity provides more robust, generator-agnostic cues than the direct utilization of raw spectral features.

\noindent\textbf{Impact of the Number of Fractal Units.} The influence of the number of Fractal Units $N$ is further investigated. Although the detection accuracy exhibits some fluctuations for specific generators, deeper fractal structures generally yield superior overall performance. Even though some models are a bit sensitive at $N=3$, the setting with $N=4$ gives the best average results across all 16 test sets. This result demonstrates that calculating multi-level fractal self-similarity can effectively improve the performance of Fractal-CNN. Therefore, the architecture with $N=4$ is identified as the optimal balance and is adopted in the final model.

\begin{figure*}
\centering
\includegraphics[width=\linewidth]{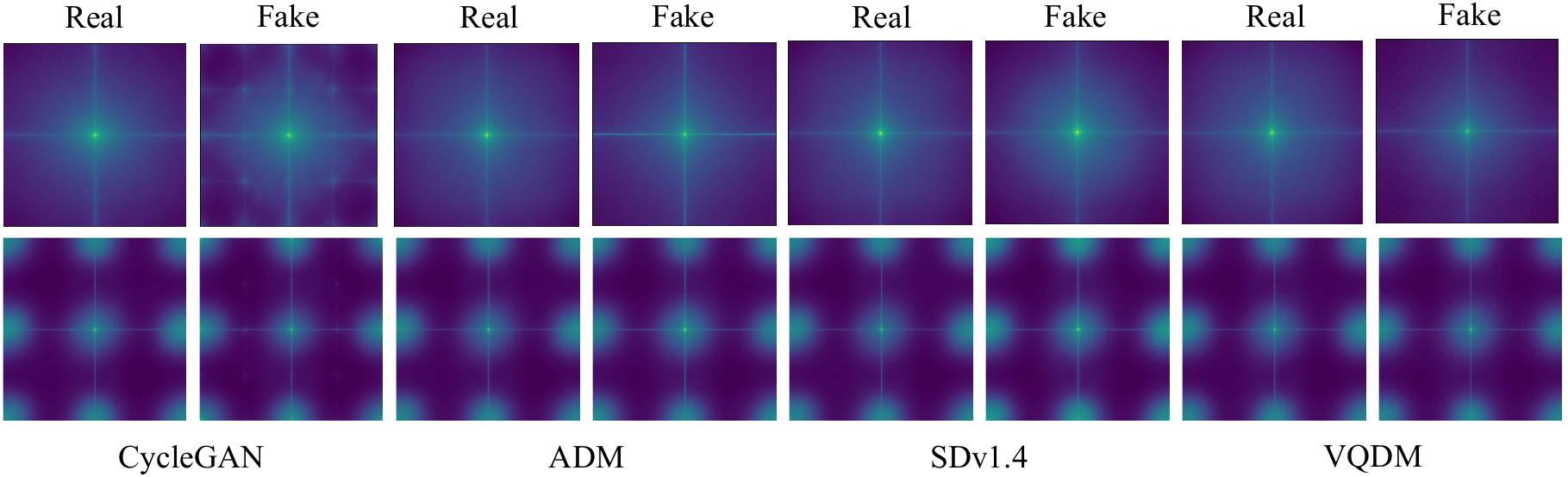}
\caption{Illustration of the spectral fractal structure. The first row displays the average spectrum of the input images $I$, and the second row shows the corresponding spectrum of the feature maps $\hat{h}$ from the backbone output.}
\label{fig:filted}
\end{figure*}

\begin{figure}
    \centering
    \includegraphics[width=\linewidth]{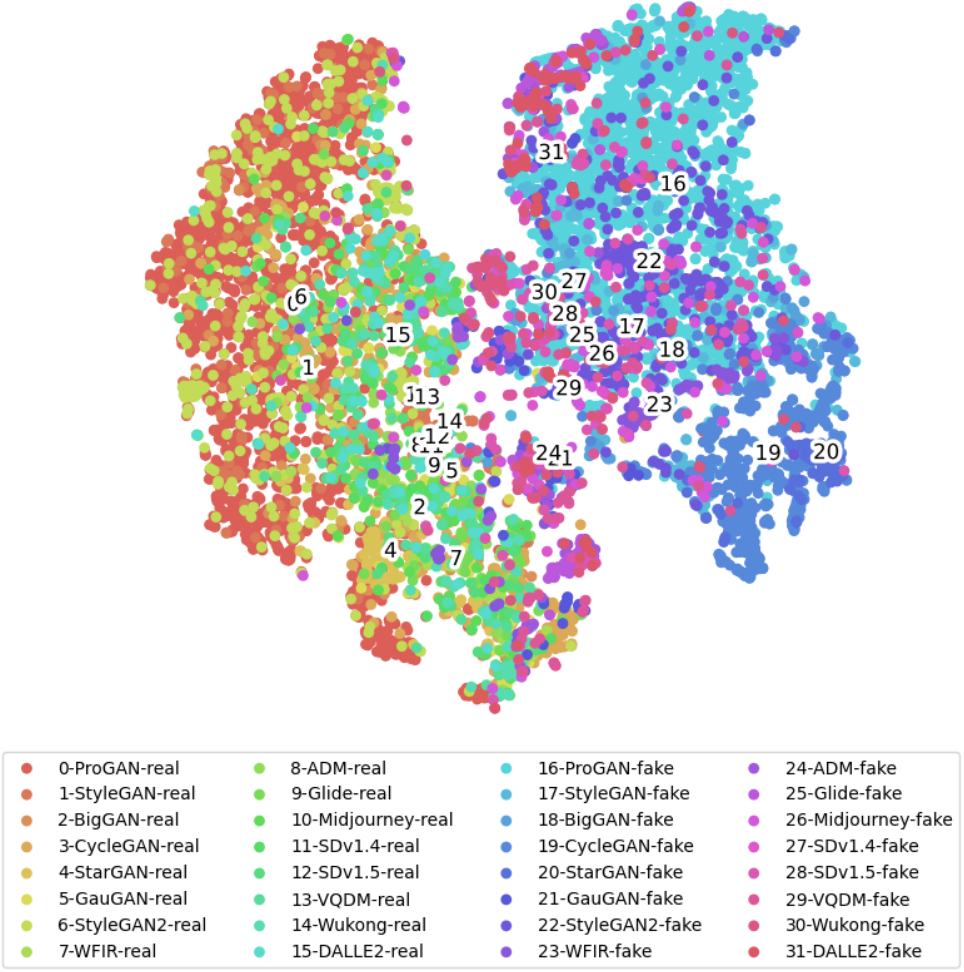}
    \caption{T-SNE visualization of the fractal self-similarity feature $S$ of real images and AI-generated images from different generators.}
    \label{fig:t-sne}
\end{figure}

\subsection{Visualizations and Discussions}

\noindent\textbf{Spectrum Visualization.} To provide an intuitive illustration of the spectral fractal structures, the magnitude spectrum of the feature map $\hat{h}$ learned by the backbone are visualized and compared to these of the original images. As depicted in \cref{fig:filted}, the spectrums are averaged over 100 images across all the channels. This analysis reveals that Fractal-CNN  primarily relies on the self-similar structure of the magnitude spectrum. Note that the grid-like frequency artifacts are not significantly amplified within the learned representation, which distinguishes the proposed approach from previous detection methods that rely heavily on such artifacts. By focusing on generator-agnostic spectral characteristics rather than generator-specific artifacts, the proposed method maintains robust generalization performance on diffusion models (e.g., ADM) where spectral artifacts are typically subtle. The consistent self-similarity observed in both GANs and Diffusion models justifies the design of our Fractal-CNN. 

\noindent\textbf{T-SNE Visualization.} To investigate the learned feature distributions, the representations are visualized through t-SNE. As illustrated in \cref{fig:t-sne}, samples from most generators form distinct clusters which remain clearly separated from the real images. This distribution demonstrates that our Fractal-CNN effectively captures consistent fractal features across diverse generative models. The internal variations within generative clusters are attributed to the noise strength $\epsilon_n$ and the mapping complexity $\mathcal{H}_n$ in \cref{eq:fractal-decomposition}. Specifically, advanced generators introduce complex mappings $\mathcal{H}_n$ which induces spectral structures with higher similarity to real images. Besides, the distribution of real images also exhibits variations across datasets. For ADM, DALLE2, and Stable Diffusion, the feature distributions of real samples show higher proximity to the AI-generated clusters. This proximity occurs because the real images in these datasets are stored in JPEG format. According to the proposed fractal model, JPEG compression acts as a compression-regeneration process which produces spectral signatures similar to these of the generative models. This observation remains consistent with the derivation in \cref{thm:fractal}, which supports the experimental results. While our Fractal-CNN captures consistent characteristics across diverse models, the results indicate that superior robustness against JPEG compression may require fine-grained source attribution.

\begin{figure*}
\centering
\includegraphics[width=0.9\linewidth]{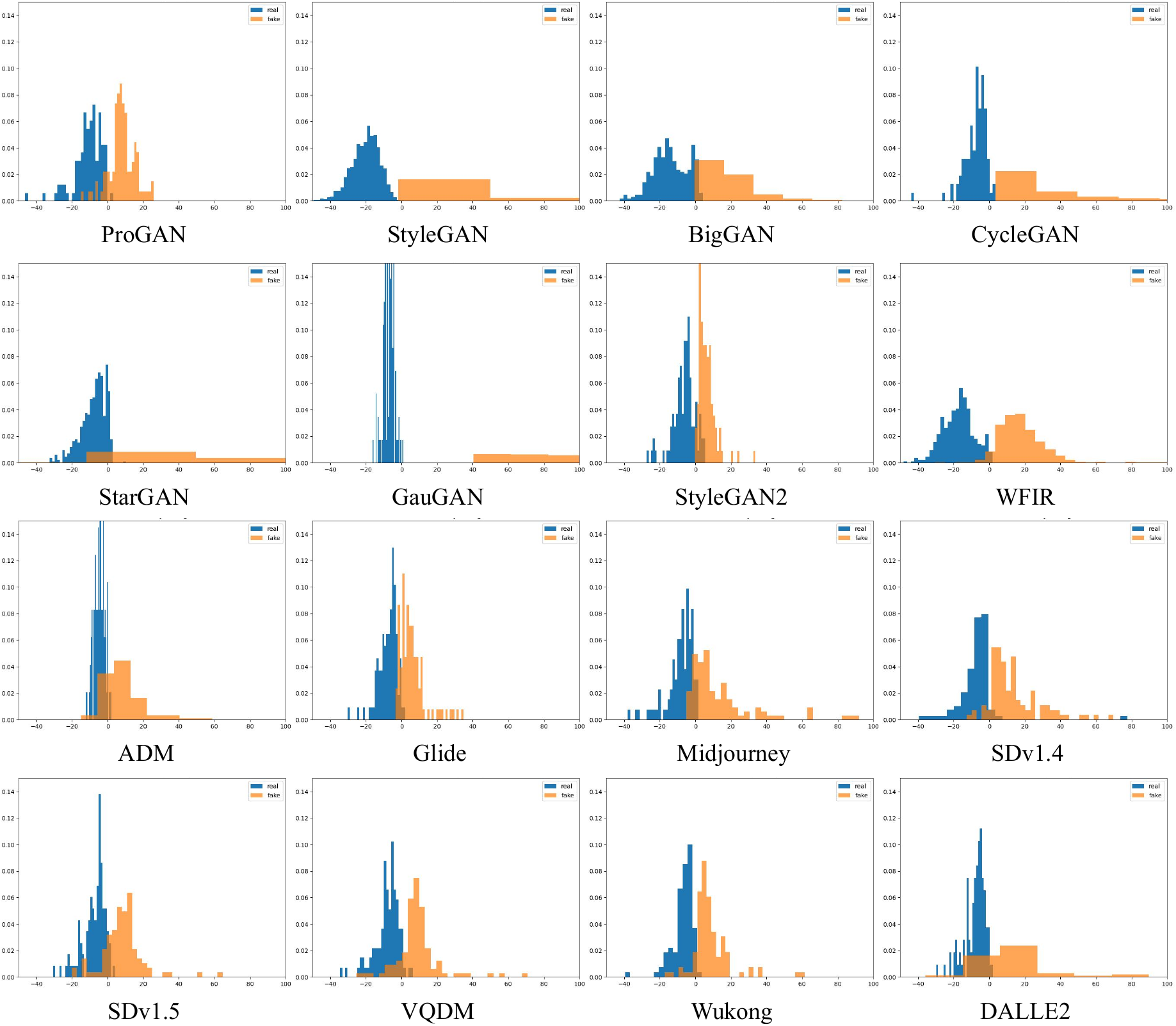}
\caption{Histogram visualization of the logit distribution from the fully connected layers. The distributions demonstrate the consistency of the learned fractal features across diverse generators.}
\label{fig:logit_hist}
\end{figure*}

\noindent\textbf{Logit Distribution Visualization. } To further evaluate how the consistent representations observed in the t-SNE visualization translate into classification stability, the histograms of the logit predictions from the final fully-connected layer are visualized. As shown in \cref{fig:logit_hist}, the logit distributions for the majority types of the generated images exhibit a substantial overlap with the distribution of the training generator (ProGAN). This observation indicates that the learned representation captures consistent fractal characteristics across diverse generators, which aligns with the clustering behavior in the feature space. Note that StyleGAN, StarGAN, and GauGAN exhibit stronger logit responses than the training distribution, which suggests a larger margin from the decision boundary. Such behavior implies that fractal self-similarity provides a stable and generator-agnostic cue to detect AI-generated images. This consistency facilitates strong generalization capabilities beyond the specific training generator, even when the spectral structures of real images undergo shifts due to JPEG compression, as previously discussed.

\section{Conclusion}

In this paper, we investigated generalizable AI-generated image detection across diverse generative models. We theoretically analyzed the spectral structure induced by dimension-increasing shift-equivariant transformations, which commonly arise in the image generation process. Our analysis shows that such transformations give rise to a generator-agnostic fractal self-similarity structure in the Fourier spectrum. Equivalently, corresponding spectral sub-regions originate from a shared lower-dimensional spectrum, leading to inherent cross-region self-similarity. Motivated by this insight, we proposed Fractal-CNN, a simple convolutional framework which captures the invariant structural self-similarity embedded in the spectrum of AI-generated images. By leveraging such model-agnostic structural cues rather than relying on raw feature patterns, the proposed method achieves improved generalization to unseen generators. Extensive experiments on both GAN- and diffusion-based models validated the effectiveness and robustness of our approach. Overall, this work highlights the importance of exploiting intrinsic structural properties of AI-generated images for generalizable detection and provides a new perspective for developing generalizable forensic methods in the face of rapidly evolving generative techniques.

\bibliographystyle{IEEEtran}
\bibliography{IEEEabrv,main}

\end{document}